\documentclass[lettersize,journal]{IEEEtran}
\usepackage{amsmath,amsfonts}

\usepackage{array}
\usepackage[caption=false,font=normalsize,labelfont=sf,textfont=sf]{subfig}
\usepackage{textcomp}
\usepackage{stfloats}
\usepackage{url}
\usepackage{verbatim}
\usepackage{graphicx}
\usepackage{cite}
\usepackage[ruled,vlined]{algorithm2e}
\usepackage{multirow}
\usepackage{booktabs}
\usepackage{array}
\usepackage{graphicx}
\usepackage{pifont}
\usepackage{amsmath}

\usepackage{hyperref}
\hypersetup{hypertex=true,
colorlinks=true,
linkcolor=blue,
anchorcolor=blue,
citecolor=blue}
\begin{document}

\title{The Essence of Balance for Self-Improving Agents in Vision-and-Language Navigation}

\author{Zhen Liu, Yuhan Liu, Jinjun Wang, Jianyi Liu, Wei Song, and Jingwen Fu%
\thanks{Z. Liu, Y. Liu, J. Wang, and J. Liu are with the State Key Laboratory of Human-Machine Hybrid Augmented Intelligence and the Institute of Artificial Intelligence and Robotics, Xi'an Jiaotong University, Xi'an 710049, China. E-mail: zhenliu9773@gmail.com. Corresponding author: J. Liu, e-mail: jyliu@xjtu.edu.cn.}%
\thanks{W. Song is with the School of Information Science and Technology, North China University of Technology, Beijing 100144, China.}%
\thanks{J. Fu is with Zhongguancun Academy, Beijing 100190, China. Project leader. E-mail: fujingwen@bza.edu.cn.}%
}

\markboth{Journal of \LaTeX\ Class Files,~Vol.~14, No.~8, August~2021}%
{Shell \MakeLowercase{\textit{et al.}}: A Sample Article Using IEEEtran.cls for IEEE Journals}

\IEEEpubid{0000--0000/00\$00.00~\copyright~2021 IEEE}

\maketitle

\begin{abstract}

In vision-and-language navigation (VLN), self-improvement from policy-induced experience—using only standard VLN action supervision—critically depends on balancing behavioral diversity and learning stability, which governs whether the agent can extract a reliable learning signal for improvement.
Increasing behavioral diversity is necessary to expose alternative action hypotheses but can destabilize policy-induced learning signals, whereas overly conservative stability constraints suppress exploration and induce early commitment, making reliable self-improvement difficult.
To address this challenge, we propose Stability–Diversity Balance (SDB), a plug-and-play mechanism for balanced self-improvement in VLN.
SDB expands each decision step into multiple latent behavioral hypotheses by applying controlled shifts in the instruction-conditioned hidden states, and then performs reliability-aware soft evaluation and aggregation to retain diverse yet instruction-consistent alternatives during learning.
An explicit regularizer further constrains hypothesis interactions, preventing excessive drift or premature collapse of hypothesis diversity and stabilizing self-improvement without discarding training signals.
Experiments on R2R, SOON, and REVERIE show consistent improvements; for example, on REVERIE val\_unseen, SDB improves SPL from 33.73 to 35.93 and OSR from 51.07 to 54.25.

\end{abstract}


\section{Introduction}

\IEEEPARstart{S}{elf-improvement}, the process by which an agent improves its competence without additional human intervention beyond standard task supervision, remains a central ambition of embodied intelligence~\cite{thrun1995lifelong,wang2016l2rl}.
In VLN, the agent operates on streaming visual observations and can be viewed as a vision-conditioned, long-horizon sequential decision problem, where temporally unstable decisions may accumulate and degrade downstream learning.
This temporal stability concern echoes a recurring theme in vision sequence modeling, where temporal-consistency learning and memory-based propagation are used to stabilize sequential predictions~\cite{liu2023tcnet,yi2020mtudm,fan2022glcor,zhou2022flowedge,zhu2022separable}.
Consequently, injecting diversity is often necessary to keep multiple plausible interpretations and action plans alive under ambiguous instructions, instead of committing too early to a single hypothesis~\cite{ke2019tactical}.
However, when diverse behaviors coexist, the agent must decide which candidate to follow and how to turn the resulting experience into stable, cumulative improvement, rather than oscillating among alternatives or collapsing diversity prematurely~\cite{lin2025evolvenav,senanayake2024predictive}.
This tension can also be interpreted as a temporal decision-consistency problem: maintaining multiple hypotheses helps cover ambiguous modes, while stable consolidation prevents temporally inconsistent switching that undermines long-horizon learning.
This raises a key question: under standard VLN action supervision, what mechanism can explicitly balance behavioral diversity and learning stability to enable dependable self-improvement in VLN?

\begin{figure}[t]
    \centering
    \includegraphics[width=1\linewidth]{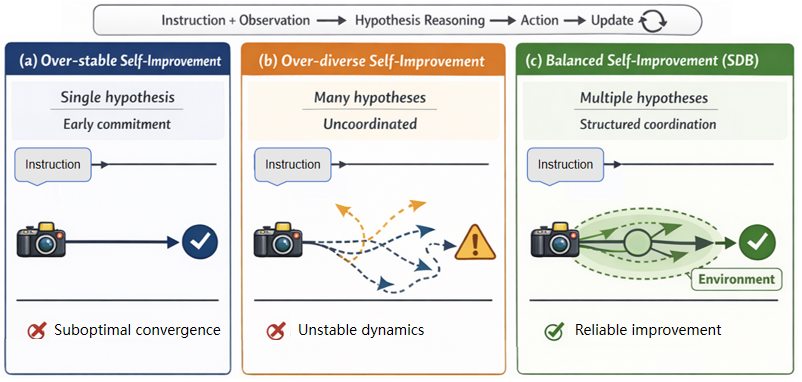}
\caption{
Stability--diversity trade-off in VLN self-improvement: over-stable learning commits to a single hypothesis, whereas over-diverse learning yields uncoordinated hypotheses and inconsistent updates; SDB encourages structured coordination.
}
    \label{fig:motivation}
\end{figure}

\IEEEpubidadjcol
To answer this question, we argue that dependable VLN self-improvement can be naturally viewed as a stability--diversity balancing problem~\cite{senanayake2024predictive,bettini2023system,lin2024curse}.
On the one hand, the agent benefits from behavioral diversity---maintaining multiple plausible latent hypotheses for the next action under linguistic ambiguity and partial observability.
On the other hand, self-improvement from policy-induced trajectories requires learning stability, so that updates remain consistent enough to accumulate progress across iterations.
This creates an inherent tension: increasing diversity can uncover better hypotheses under ambiguity, but may introduce inefficient exploration and competing objectives; in contrast, overly conservative stability encourages early commitment and can trap the agent in suboptimal local optima.
Fig.~\ref{fig:motivation} summarizes this trade-off and the two failure patterns at either extreme.
However, a number of VLN systems~\cite{gu2022vlnsurvey,wu2021vlnsurvey} handle diversity and reliability in a disconnected manner: diversity is often introduced post-hoc at inference (e.g., backtracking/search~\cite{fried2018speaker,ke2019tactical,ma2019regretful} or multi-hypothesis tracking~\cite{anderson2019chasing}), while reliability is enforced via heuristic monitors or selection rules~\cite{ma2019regretful}.
As a result, the training-time interaction between diversity and stability is often left implicit, rather than being explicitly modeled and optimized as a coupled objective.

Prior VLN research has explored several directions to cope with partial observability, long-horizon planning, and instruction ambiguity.
A line of work improves state inference by incorporating history or memory, aiming to make decisions less myopic under incomplete observations~\cite{chen2021history,krantz2020beyond}.
Another line targets goal grounding beyond the current viewpoint by explicitly reasoning about unseen landmarks and remote goals, as exemplified by object- or region-centric navigation benchmarks and methods~\cite{qi2020reverie,ku2020rxr}.
To handle ambiguity and recover from wrong turns, many systems introduce behavioral diversity via backtracking, regretful reasoning, or multi-path exploration at inference time, effectively maintaining multiple candidate routes or hypotheses~\cite{ma2019regretful,li2019robust}.
Meanwhile, exploration-oriented techniques such as parameter-space noise and stochastic action perturbations have been explored to promote behavioral exploration in sequential decision making~\cite{plappert2018paramnoise,fortunato2018noisynet}.
Despite these advances, diversity and stability are typically addressed by separate mechanisms (often at different stages), and the training objective rarely treats their interaction as a coupled quantity to be explicitly optimized, especially when learning relies on policy-induced trajectories~\cite{wang2019reinforced,anderson2018evaluation,lin2024curse}. 

This motivates a training-time (1$\rightarrow$K$\rightarrow$1) expand--select design: a Diversity Expansion Module to generate multiple plausible hypotheses, and a Stability Selection Module to score and consolidate them for both learning and execution.
To address this gap, we propose Stability--Diversity Balance (SDB), a plug-in expand--select mechanism that explicitly couples stability and diversity during training for vision-conditioned sequential decision making.
SDB follows a structured (1$\rightarrow$K$\rightarrow$1) procedure: starting from a single decision state, it generates $K$ latent behavioral hypotheses via controlled hidden-state perturbations, evaluates their reliability, and then softly consolidates them into a coordinated update under an SDB regularizer.
This design preserves informative diversity for resolving ambiguity while maintaining stable, cumulative learning dynamics across iterations, rather than leaving hypothesis interaction unconstrained~\cite{anderson2019chasing,lin2024curse}.
As a result, SDB enables more dependable self-improvement from policy-induced trajectories without requiring additional human supervision beyond the standard action labels used in VLN training.

We evaluate SDB on standard VLN benchmarks, including R2R~\cite{anderson2018r2r}, SOON~\cite{zhu2021soon}, and REVERIE~\cite{qi2020reverie}, instantiated on three representative backbones: DUET~\cite{chen2022think}, GOAT~\cite{wang2024goat}, and the VLM-based NavGPT-2~\cite{zhou2024navgpt2}.
Across benchmarks and model backbones, integrating SDB generally improves both navigation success and path efficiency over the corresponding baselines.
On the challenging REVERIE val\_unseen split, DUET+SDB increases SPL from 33.73 to 35.93.
Overall, these results support that making the stability--diversity interaction explicit during training can improve reliability and efficiency when learning from policy-induced trajectories in VLN.

Our contributions are summarized as follows:
\begin{itemize}
    \item We formulate VLN self-improvement under standard action supervision through a stability--diversity lens, and identify two common failure patterns at the extremes: early over-commitment to a single hypothesis versus insufficient coordination among diverse hypotheses.
    \item We introduce Stability--Diversity Balance (SDB), a backbone-agnostic, training-time (1$\rightarrow$K$\rightarrow$1) expand--select plug-in that comprises a Diversity Expansion Module and a Stability Selection Module, explicitly coupling diversity and stability via an SDB regularizer and reliability-aware consolidation.
    \item We demonstrate the effectiveness of SDB on R2R, SOON, and REVERIE with three representative backbones (DUET, GOAT, and the VLM-based NavGPT-2), achieving improved navigation success and efficiency over strong baselines.
\end{itemize}

\section{Related Work}

\subsection{Self-improvement in Embodied Intelligence}
Improving an embodied agent from its own interaction experience is a long-standing goal, spanning continual/lifelong learning and meta-learning that update policies through accumulated interaction and fast adaptation~\cite{parisi2019continual,hospedales2022metalearning,finn2017maml}.
In sequential decision-making, self-improvement is challenging because learning is shaped by policy-induced trajectories: early suboptimal decisions can steer future visitation and complicate the attribution of credit, especially when supervision is derived from on-policy rollouts~\cite{ross2011dagger,bengio2015scheduled}.
This has motivated interactive imitation learning and stabilized policy optimization strategies that aim to leverage alternative behaviors while maintaining stable training dynamics~\cite{ross2014interactive,schulman2017ppo}.

\subsection{Diversity--Stability Trade-off in Agent Learning}
A common way to increase behavioral diversity is to maintain multiple candidates, for example via ensembles for exploration~\cite{osband2016bootstrapped}.
Complementary to action-space noise, parameter-space noise and learned parametric noise (NoisyNets) perturb internal representations to induce consistent exploratory behaviors~\cite{plappert2018paramnoise,fortunato2018noisynet}, conceptually related to generating multiple latent hypotheses by shifting hidden states.
However, diversity is not always beneficial: recent evidence shows that overly diverse ensembles can introduce harmful interference and degrade learning, known as the ``curse of diversity''~\cite{lin2024curse}.
Related stability issues have also been extensively studied in vision sequence modeling~\cite{liu2023tcnet,yi2020mtudm,gui2020bilateral,fan2022glcor,zhou2022flowedge,zhu2022separable}, where temporally inconsistent predictions can accumulate and degrade long-horizon quality.
To address this, prior work often introduces temporal-consistency constraints, correspondence learning, or memory-based propagation to stabilize sequential outputs.
These findings provide complementary evidence that stability should be treated as an explicit objective when leveraging multiple hypotheses in sequential decision making~\cite{bettini2023system,senanayake2024predictive}.

\subsection{Vision-and-Language Navigation}
Vision-and-Language Navigation (VLN) requires an agent to follow natural-language instructions in photorealistic environments~\cite{anderson2018r2r}, and modern benchmarks such as REVERIE and SOON further increase object grounding and linguistic diversity~\cite{qi2020reverie,zhu2021soon}.
To address ambiguity and generalization, prior models leverage richer history-aware representations~\cite{chen2021history,qiao2023hop+} and stronger structural biases (e.g., DUET)~\cite{chen2022duet}, while complementary efforts improve robustness to spurious correlations (e.g., GOAT)~\cite{wang2024goat}.
In parallel, large vision-language models have been integrated for explicit reasoning (e.g., NavGPT-2~\cite{zhou2024navgpt2}), bridging plan generation and action prediction.

From a sequential perspective, VLN is long-horizon and partially observable: small early ambiguities can lead to divergent trajectories, and decision instability may manifest as frequent plan switching, detours, or inconsistent updates when learning from policy-induced trajectories~\cite{anderson2019chasing,lin2024curse}.
This motivates training-time mechanisms that can maintain multiple plausible hypotheses under ambiguity while stabilizing how these alternatives are consolidated into learning signals.
We address this need with SDB, a lightweight and backbone-agnostic plug-in applicable to both specialist VLN backbones and VLM-based agents.

\begin{figure*}[t]
    \centering
    \includegraphics[width=\textwidth]{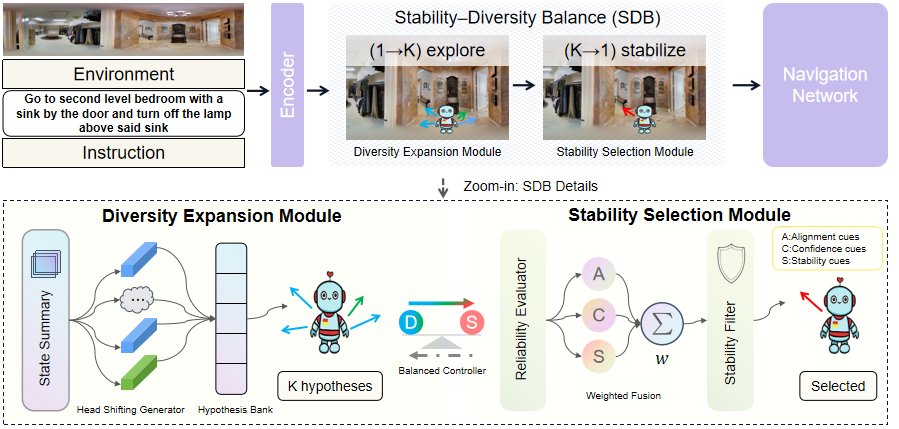}
\caption{\textbf{Overview of Stability--Diversity Balance (SDB).}
SDB is inserted between the backbone encoder and the navigation policy network.
It expands the decision context into $K$ hypotheses (1$\rightarrow$K, explore) and consolidates them into a single reliable decision representation (K$\rightarrow$1, stabilize) for training and execution.}
    \label{fig:sdb_overview}
\end{figure*}

\section{Method}
\label{sec:method}

We study VLN under standard action-label supervision and aim to improve from policy-induced trajectories by leveraging the agent's internal hypothesis space.
Here, ``self-improvement'' refers to improving decision making by exploiting internally generated hypotheses under standard action supervision,
without requiring additional human annotations or extra environment data collection.
A core obstacle is the stability--diversity tension: increasing behavioral diversity can resolve ambiguity under partial observability,
yet uncontrolled diversity may induce inconsistent internal decisions and destabilize learning updates; overly conservative stabilization, in contrast, suppresses exploration and induces early commitment.

We propose {Stability--Diversity Balance (SDB)}, which implements balance as a structured training-time $(1\!\rightarrow\!K\!\rightarrow\!1)$ decision operator inserted between the backbone encoder and the navigation policy network.
Importantly, SDB is not a generic ensembling heuristic or stochastic regularizer: it factorizes hypothesis expansion (diversity) and reliability-aware commitment (stability) into explicit, attributable steps, enabling coordinated credit assignment under policy-induced trajectories.
SDB keeps the backbone architecture unchanged and only acts on the encoder-produced token streams $(\mathbf{T}_t,\mathbf{E}_t)$ via a lightweight hypothesis module.
Accordingly, SDB is neither an inference-time search mechanism (e.g., beam/backtracking or multi-particle tracking) nor an unstructured perturbation scheme (e.g., dropout or parameter noise).
Instead, it adopts a factorized and controllable hypothesis parameterization: diversity is generated through bounded, evidence-conditioned low-rank hidden-state shifts with explicit slot gating,
while stability is learned via reliability-aware consolidation driven by explicit Alignment/Confidence/Stability (ACS) cues.
This design makes hypothesis generation attributable (via shift bases and slot gating) and stabilization coordinated (via reliability-weighted fusion and an explicit regularizer),
rather than relying on heuristic sampling-and-picking.
As illustrated in Fig.~\ref{fig:sdb_overview}, SDB expands the decision context into $K$ latent hypotheses $(1\!\rightarrow\!K)$ and then consolidates them into a single reliable representation $(K\!\rightarrow\!1)$ before action prediction.
Here $K$ includes an unshifted anchor hypothesis and is treated as a hyperparameter specified in the implementation details.
﻿

\subsection{VLN Formulation}
\label{sec:task}

VLN is defined on a navigable graph $G=(V,E)$.
Given an instruction $I$, the agent starts at $v_1$ and runs for at most $T$ steps.
At step $t$, the agent observes a panoramic view $O_t$ at viewpoint $v_t$ and predicts an action from a dynamic candidate set $\mathcal{A}_t$ consisting of navigable neighbor moves plus \textsc{Stop}.
We denote the policy distribution as $p_t(\cdot)$ over $\mathcal{A}_t$ and always include \textsc{Stop} in $\mathcal{A}_t$.

\textbf{Backbone.}
We adopt VLN-DUET as the backbone for exposition.
At each step $t$, DUET produces instruction token embeddings $\mathbf{T}_t\in\mathbb{R}^{B\times L\times H}$ and environment evidence composed of (i) viewpoint tokens and (ii) global-map tokens.
We jointly denote the environment-side evidence tokens as $\mathbf{E}_t$.
The backbone encoder yields a decision context representation, which is then fed to the navigation policy network (action head).

\subsection{SDB Overview: Balance as a Controllable Decision Operator}
\label{sec:sdb_overview}

SDB balances two coupled requirements.
\textbf{Diversity} means generating $K$ plausible latent behavioral hypotheses under the same instruction--environment evidence, so the agent can consider alternative interpretations under ambiguity.
\textbf{Stability} means committing to a temporally consistent navigation decision and producing updates that can accumulate progress across iterations.
SDB operates on the encoder-produced decision context and outputs a single stabilized latent state for the navigation policy network, leaving the backbone encoder unchanged.

At step $t$, the Diversity Expansion module produces $K$ hypothesis states from the same decision context.
The Stability Selection module evaluates hypothesis reliability and (i) produces a soft consolidated context for training, while (ii) committing to a single hypothesis for execution.
A lightweight Balanced Controller provides the reliability weights that couple expansion and selection, and also drives a small auxiliary regularizer to prevent degenerate hypothesis behaviors.

\subsection{Diversity Expansion Module (1$\rightarrow$K)}
\label{sec:diversity}

Diversity Expansion aims to generate $K$ plausible latent behavioral hypotheses under the same instruction--environment evidence, without committing to a final decision.
Following Fig.~\ref{fig:sdb_overview}, we implement it with a Head-Shifting Generator that creates a Hypothesis Bank.
Hypothesis generation is conditioned on the current step evidence and navigation stage via a compact state summary,
yielding bounded low-rank shifts that produce controllable, instruction-consistent variations rather than unconstrained deviations.

\textbf{State summary.}
Under partial observability, hypothesis generation should depend on the current evidence and navigation stage, rather than relying on uncontrolled randomness.
Let instruction tokens be $\mathbf{T}_t$ and let environment evidence be denoted by $\mathbf{E}_t$.
We summarize the step by pooling the available tokens and injecting a step embedding:
\begin{equation}
\begin{aligned}
[\bar{\mathbf{t}}_t,\bar{\mathbf{e}}_t] &= [\mathrm{Pool}(\mathbf{T}_t),\,\mathrm{Pool}(\mathbf{E}_t)],\\
\mathbf{s}_t &= \phi\!\left([\bar{\mathbf{t}}_t,\bar{\mathbf{e}}_t,\mathbf{e}(t)]\right).
\end{aligned}
\label{eq:state_summary_sdb}
\end{equation}
where $\mathrm{Pool}(\cdot)$ denotes masked mean pooling, $\mathbf{e}(t)$ is a step embedding, and $\phi(\cdot)$ is a lightweight MLP.

\textbf{Head-Shifting Generator (HSG).}
We parameterize hypothesis shifts via a shared low-rank, evidence-conditioned operator and a hypothesis-wise gating.
Let $\mathbf{s}_t$ be a compact state summary at step $t$ derived from the current instruction--environment evidence.

We first build a shared low-rank semantic shift basis
\begin{equation}
\mathbf{B}_t = \mathbf{W}_u \,\mathrm{Diag}\!\big(\sigma(\mathbf{W}_b \mathbf{s}_t)\big)\,\mathbf{W}_d ,
\label{eq:hsg_basis}
\end{equation}
where $\mathbf{W}_b$ maps $\mathbf{s}_t$ to an evidence-dependent gating vector, $\sigma(\cdot)$ (sigmoid) bounds each gate to $(0,1)$ to enforce bounded shifts, and $\mathrm{Diag}(\cdot)$ turns the gating vector into a diagonal matrix that scales shared low-rank semantic directions.
$\mathbf{W}_d$ and $\mathbf{W}_u$ are shared projection matrices that down-/up-project features through a low-rank semantic subspace, yielding an evidence-conditioned shift operator $\mathbf{B}_t$.

We then compute a state-conditioned gating over $K$ hypothesis slots
\begin{equation}
\boldsymbol{\pi}_t = \mathrm{softmax}(\mathbf{W}_\pi \mathbf{s}_t).
\label{eq:hsg_gating}
\end{equation}
Here $\boldsymbol{\pi}_t=\{\pi_t^{(k)}\}_{k=0}^{K-1}$ assigns a state-adaptive weight to each slot.
We designate $k{=}0$ as the unshifted anchor hypothesis, while slots $k\ge 1$ correspond to shifted hypotheses.

For each shifted slot $k\ge 1$, we generate a bounded shift as
\begin{equation}
\Delta \mathbf{T}_t^{(k)} =
\pi_t^{(k)}\, \mathrm{LN}(\mathbf{T}_t)\mathbf{B}_t
\;+\;
\gamma\,\pi_t^{(k)}\, \Delta \mathbf{T}_{t,\mathrm{align}}^{(k)},
\label{eq:hsg_shift}
\end{equation}
where the first term applies a global semantic shift through the shared low-rank basis,
and $\Delta \mathbf{T}_{t,\mathrm{align}}^{(k)}$ is a fine-grained instruction--environment alignment residual, which corrects slot-specific misalignment with the current evidence.
We treat $\gamma$ as a learnable scalar and parameterize it as
\begin{equation}
\gamma=\sigma(\theta_\gamma),
\label{eq:gamma_learn}
\end{equation}
to keep its contribution bounded.

In our implementation, the alignment residual is produced by a lightweight cross-attention block from instruction to environment evidence with a hypothesis-specific query bias:
\begin{equation}
\Delta \mathbf{T}_{t,\mathrm{align}}^{(k)}=
\mathrm{CA}\!\left(\mathrm{LN}(\mathbf{T}_t),\,\mathbf{E}_t;\,\delta\mathbf{q}_t^{(k)}\right)\mathbf{W}_{\mathrm{align}},
\label{eq:align_residual}
\end{equation}
where $\mathrm{CA}(\cdot)$ denotes a single cross-attention layer, $\delta\mathbf{q}_t^{(k)}$ is added to the instruction queries in cross-attention to modulate evidence access under each hypothesis, and $\mathbf{W}_{\mathrm{align}}$ is a linear projection.

The hypothesis instruction states are then
\begin{equation}
\tilde{\mathbf{T}}_t^{(k)}=
\begin{cases}
\mathbf{T}_t, & k=0,\\[2pt]
\mathrm{LN}(\mathbf{T}_t)+\Delta\mathbf{T}_t^{(k)}, & k=1,\ldots,K-1,
\end{cases}
\label{eq:hsg_shifted_txt}
\end{equation}
yielding $K$ instruction-side hypothesis states $\{\tilde{\mathbf{T}}_t^{(k)}\}_{k=0}^{K-1}$ that are diverse yet grounded in the same environment evidence.

\textbf{Hypothesis contexts and bank.}
We obtain hypothesis contexts by feeding the shifted instruction tokens and shared environment evidence into a lightweight fusion block with shared parameters:
\begin{equation}
\mathbf{H}_t^{(k)} = f_{\eta}\!\left(\tilde{\mathbf{T}}_t^{(k)}, \mathbf{E}_t; \delta\mathbf{q}_t^{(k)}\right), \qquad k=0,\ldots,K-1,
\label{eq:hyp_context_def}
\end{equation}
where $f_{\eta}$ consists of a single instruction-to-environment cross-attention layer followed by a feed-forward layer, and is shared across hypotheses.
We store the $K$ hypothesis contexts as a hypothesis bank
\begin{equation}
\mathcal{H}_t=\left\{\mathbf{H}_t^{(k)}\right\}_{k=0}^{K-1}.
\label{eq:hypothesis_bank_sdb}
\end{equation}
All hypotheses share the same environment evidence; diversity arises from the HSG-induced shifts and alignment modulation.
The bank $\mathcal{H}_t$ is passed to the Stability Selection module for consolidation and selection.

\subsection{Stability Selection Module ($K\!\rightarrow\!1$)}
\label{sec:stability}

Given the hypothesis bank $\{\mathbf{H}_t^{(k)}\}_{k=0}^{K-1}$ at step $t$, the stability selection module produces a reliable decision context by
(i) evaluating hypothesis reliability, (ii) consolidating hypotheses for training via weighted fusion, and (iii) committing to a single hypothesis for execution via stable selection.
The resulting reliability weights thus couple learning and execution: they drive soft consolidation during training to provide smoother gradients,
and enable temporally stable commitment during inference to reduce oscillation.

\textbf{Reliability evaluator (ACS cues).}
Each hypothesis $k$ is scored using three groups of lightweight cues aligned with Fig.~\ref{fig:sdb_overview}:
Alignment cues (A) reflect how consistent a hypothesis remains with the anchor under instruction-aligned representations;
Confidence cues (C) reflect local decisiveness/uncertainty from environment-conditioned representations; and
Stability cues (S) reflect temporally consistent navigation potential.
We compute cues from pooled hypothesis descriptors $\bar{\mathbf{h}}_t^{(k)}=\mathrm{Pool}(\mathbf{H}_t^{(k)})$:
\begin{equation}
\begin{aligned}
A_t^{(k)} &= \cos\!\big(\bar{\mathbf{h}}_t^{(k)},\bar{\mathbf{h}}_t^{(0)}\big),\\
C_t^{(k)} &= -\mathcal{H}\!\big(p_t^{(k)}\big),\\
S_t^{(k)} &= \cos\!\big(\bar{\mathbf{h}}_t^{(k)},\bar{\mathbf{h}}_{t-1}^{\dagger}\big).
\end{aligned}
\label{eq:acs_cues}
\end{equation}
where $p_t^{(k)}$ denotes the action distribution predicted from $\mathbf{H}_t^{(k)}$ by the shared navigation head, $\mathcal{H}(\cdot)$ is entropy, and $\bar{\mathbf{h}}_{t-1}^{\dagger}$ is the pooled descriptor of the context used at step $t{-}1$ (defined as $\mathrm{Pool}(\mathbf{H}_{t-1}^{\mathrm{acs}})$ during training and $\mathrm{Pool}(\mathbf{H}_{t-1}^{*})$ during execution; initialized at $t{=}1$ with the anchor).
We concatenate cues into $\mathbf{r}_t^{(k)}=[A_t^{(k)},C_t^{(k)},S_t^{(k)}]$ and map it to a scalar score via a small MLP:
\begin{equation}
a_t^{(k)} = g_{\psi}\!\left(\mathbf{r}_t^{(k)}\right), \qquad k=0,\ldots,K-1,
\label{eq:acs_score}
\end{equation}
where $k=0$ denotes the anchor hypothesis.
Computing $\{p_t^{(k)}\}$ incurs an additional cost that scales linearly with $K$, but remains lightweight in practice due to the small hypothesis module and shared head.

\textbf{Weighted fusion (ACS and $\mathbf{w}$).}
The Balanced Controller converts the reliability scores into normalized weights by softmax over hypotheses:
\begin{equation}
w_t^{(k)}=\mathrm{softmax}_k\!\left(a_t^{(k)}\right), 
\qquad
\mathbf{w}_t=\{w_t^{(k)}\}_{k=0}^{K-1}.
\label{eq:acs_weight}
\end{equation}
During training, we use $\mathbf{w}_t$ to form a soft consolidated context
\begin{equation}
\mathbf{H}_t^{\mathrm{acs}}=\sum_{k=0}^{K-1} w_t^{(k)}\,\mathbf{H}_t^{(k)},
\label{eq:acs_fusion_compact}
\end{equation}
which provides smoother gradients and reduces sensitivity to hypothesis switching.

\textbf{Stability select.}
During execution, we commit to a single hypothesis to reduce oscillations and encourage temporally consistent actions.
We perform a lightweight temporal smoothing on the controller weights using an exponential moving average (EMA):
\begin{equation}
\begin{aligned}
\bar{\mathbf{w}}_t &= (1-\rho)\bar{\mathbf{w}}_{t-1}+\rho \mathbf{w}_t,\\
k_t^{*} &= \arg\max_{k}~\bar{w}_t^{(k)},\\
\mathbf{H}_t^{*} &= \mathbf{H}_t^{(k_t^{*})}.
\end{aligned}
\label{eq:stable_select}
\end{equation}
We initialize $\bar{\mathbf{w}}_1=\mathbf{w}_1$.
We treat $\rho$ as a learnable scalar and parameterize it as
\begin{equation}
\rho=\sigma(\theta_\rho),
\label{eq:rho_learn}
\end{equation}
to keep $\rho\in(0,1)$ and avoid degenerate smoothing.
In short, SDB uses soft consolidation for training (Eq.~\ref{eq:acs_fusion_compact}) and stabilized commitment for execution (Eq.~\ref{eq:stable_select}), matching the weighted-fusion and stability-filter design in Fig.~\ref{fig:sdb_overview}.

\subsection{Training Objective and Reliability-Aware Self-Improvement}
\label{sec:train}

\textbf{DUET-style supervised objective.}
We follow standard DUET-style training and optimize action prediction with cross-entropy over the candidate action set at each step:
\begin{equation}
\mathcal{L}_{\mathrm{duet}}
=
-\sum_t \log p_t(a_t^\ast \mid O_t,I),
\label{eq:duet_loss}
\end{equation}
where $a_t^\ast$ is the ground-truth action and $p_t(\cdot)$ is produced by the navigation policy network.
(When an auxiliary head is available, e.g., object grounding in REVERIE, we add the corresponding supervised term in the same form.)

\textbf{Reliability-aware training signal.}
At each step, SDB generates $K$ hypotheses and the Balanced Controller outputs weights $\mathbf{w}_t$ (Eq.~\ref{eq:acs_weight}).
During training, we feed the soft consolidated context $\mathbf{H}_t^{\mathrm{acs}}$ (Eq.~\ref{eq:acs_fusion_compact}) to the navigation network, so learning benefits from the model's own hypothesis space under standard action-label supervision, without introducing extra supervision or collecting additional on-policy trajectories.

\textbf{SDB regularizer $\mathcal{L}_{\mathrm{sdb}}$.}
To keep the hypothesis dynamics diverse but coordinated, we add a lightweight regularizer computed from hypothesis representations and controller weights.
Let $\bar{\mathbf{h}}_t^{(k)}=\mathrm{Pool}(\mathbf{H}_t^{(k)})$ and $\bar{\mathbf{h}}_t^{\mathrm{acs}}=\mathrm{Pool}(\mathbf{H}_t^{\mathrm{acs}})$.
We instantiate three components:
\begin{equation}
\begin{aligned}
\mathcal{L}_{\mathrm{agr}}
&=
\sum_{k=0}^{K-1} w_t^{(k)}
\left\lVert \bar{\mathbf{h}}_t^{(k)}-\bar{\mathbf{h}}_t^{\mathrm{acs}} \right\rVert_2^2,\\
\mathcal{L}_{\mathrm{sm}}
&=
\sum_{k=1}^{K-1}
\left\lVert \bar{\mathbf{h}}_t^{(k)}-\bar{\mathbf{h}}_t^{(k-1)} \right\rVert_2^2,\\
\mathcal{L}_{\mathrm{div}}
&=
\max\!\left(0,\,m-\mathrm{Var}_{k}(\bar{\mathbf{h}}_t^{(k)})\right).
\end{aligned}
\label{eq:l_agr_l_sm}
\end{equation}
where $\mathcal{L}_{\mathrm{agr}}$ enforces reliability-weighted coordination by pulling hypotheses toward the consolidated representation,
$\mathcal{L}_{\mathrm{sm}}$ penalizes abrupt changes across neighboring slots (as induced by the ordered shift/gating in Eq.~\ref{eq:hsg_gating}),
and $\mathcal{L}_{\mathrm{div}}$ enforces a minimum diversity floor.
We define $\mathrm{Var}_{k}(\bar{\mathbf{h}}_t^{(k)})$ as the mean per-dimension variance across hypotheses, i.e., $\frac{1}{H}\sum_{j=1}^{H}\mathrm{Var}_{k}(\bar{\mathbf{h}}_{t,j}^{(k)})$.
We treat $m$ as a learnable non-negative scalar and parameterize it as
\begin{equation}
m=\mathrm{softplus}(\theta_m).
\label{eq:m_learn}
\end{equation}

We write the regularizer compactly as
\begin{equation}
\mathcal{L}_{\mathrm{sdb}}
=
\lambda_{\mathrm{agr}}\mathcal{L}_{\mathrm{agr}}
+
\lambda_{\mathrm{sm}}\mathcal{L}_{\mathrm{sm}}
+
\lambda_{\mathrm{div}}\mathcal{L}_{\mathrm{div}},
\label{eq:lsdb_compact}
\end{equation}
where all components are computed from pooled hypothesis descriptors and controller weights, and $\lambda$'s are small constants.

\textbf{Total objective.}
We combine the supervised DUET loss and the SDB regularizer with a learnable non-negative weight:
\begin{equation}
\omega=\mathrm{softplus}(\theta), \qquad
\mathcal{L}_{\mathrm{train}}=\mathcal{L}_{\mathrm{duet}}+\omega\,\mathcal{L}_{\mathrm{sdb}}.
\label{eq:train_objective}
\end{equation}
This keeps the main learning signal aligned with standard DUET-style supervision, while allowing the model to adaptively calibrate how strongly hypothesis regularization should shape stable self-improvement.
All learnable scalars ($\rho$, $\gamma$, $m$, and $\omega$) are initialized to non-degenerate starting values and optimized jointly with the network (implementation details).

\section{Experiments}
To evaluate whether Stability--Diversity Balance (SDB) yields consistent benefits under standard VLN supervision, we conduct experiments on three established benchmarks: R2R, SOON, and REVERIE.
These datasets cover step-by-step instruction following (R2R), long-horizon navigation with higher linguistic complexity (SOON), and navigation with object grounding requirements (REVERIE).
We follow the official splits and evaluation scripts, and report standard metrics for fair, protocol-aligned comparisons~\cite{anderson2018vision,fried2018speaker}.

\subsection{Datasets}
\textbf{R2R}~\cite{anderson2018vision} is a classic VLN benchmark with relatively short trajectories and step-by-step instructions, focusing on instruction following and generalization to unseen environments.

\textbf{SOON}~\cite{zhu2021soon} features longer, more challenging navigation episodes and instructions that require stronger contextual reasoning, stressing long-horizon planning and robustness.

\textbf{REVERIE}~\cite{qi2020reverie} extends VLN with object grounding: beyond reaching the target location, the agent must stop and localize an object described in the instruction, emphasizing fine-grained vision--language alignment in complex scenes.

\subsection{Evaluation Metrics}
\label{sec:metrics}
We follow the official VLN evaluation protocols and report metrics that jointly measure task success and navigation efficiency.
\textbf{Trajectory Length (TL)} measures the average executed path length (lower is better).
\textbf{Success Rate (SR)} is the proportion of episodes in which the agent stops within a threshold distance $\delta$ of the goal.
\textbf{Oracle Success Rate (OSR)} measures whether the agent ever comes within $\delta$ of the goal at any step (regardless of where it stops).
\textbf{Success weighted by Path Length (SPL)} jointly evaluates success and efficiency by rewarding successful episodes that follow shorter paths.
For \textbf{R2R}, we additionally report \textbf{Navigation Error (NE)} as standard.
Unless otherwise specified, we use the standard success threshold $\delta$ adopted by each benchmark.

\begin{table*}[t]
	\centering
    \footnotesize
    \setlength{\tabcolsep}{4pt}
    \renewcommand{\arraystretch}{1.08}
	\caption{Performance comparison on the REVERIE dataset across Val Seen, Val Unseen, and Test Unseen splits.}
	\label{tab:reverie_results}
	\resizebox{\textwidth}{!}{
		\begin{tabular}{lcccc|cccc|cccc}
			\toprule
			& \multicolumn{4}{c}{\textbf{Val Seen}} & \multicolumn{4}{c}{\textbf{Val Unseen}} & \multicolumn{4}{c}{\textbf{Test Unseen}} \\
			\cmidrule(lr){2-5} \cmidrule(lr){6-9} \cmidrule(lr){10-13}
			\textbf{Model} & \textbf{TL} (↓) & \textbf{OSR} (↑) & \textbf{SR} (↑) & \textbf{SPL} (↑) & \textbf{TL} (↓) & \textbf{OSR} (↑) & \textbf{SR} (↑) & \textbf{SPL} (↑) & \textbf{TL} (↓) & \textbf{OSR} (↑) & \textbf{SR} (↑) & \textbf{SPL} (↑) \\
			\midrule
			Seq2Seq \cite{anderson2018vision} & 12.88 & 35.70 & 29.59 & 24.01 & 11.07 & 8.07 & 4.20 & 2.84 & 10.89 & 6.88 & 3.99 & 3.09 \\
			SMNA \cite{ma2019selfmonitoring} & \textbf{7.54} & 43.29 & 41.25 & 39.61 & \textbf{9.07} & 11.28 & 8.15 & 6.44 & \textbf{9.23} & 8.39 & 5.80 & 4.53 \\
			RecBERT \cite{hong2021vln} & 13.44 & 53.90 & 51.79 & 47.96 & 16.78 & 35.02 & 30.67 & 24.90 & 15.86 & 32.91 & 29.61 & 23.99 \\
			HAMT \cite{chen2021history} & 12.79 & 47.65 & 43.29 & 40.19 & 14.08 & 36.84 & 32.95 & 30.20 & 13.62 & 33.41 & 30.40 & 26.67 \\
			DUET \cite{chen2022think} & 13.86 & \textbf{73.86} & \textbf{71.75} & \textbf{63.94} & 22.11 & 51.07 & 46.98 & 33.73 & 21.59 & {56.71} & {52.51} & 36.06 \\
			\textbf{DUET+SDB} & 13.83 & {70.56} & {69.50} & {62.07} & 23.22 & \textbf{54.25} & \textbf{50.81} & \textbf{35.93} & {22.35} & \textbf{58.63} & \textbf{54.92} & \textbf{37.74} \\
			\bottomrule
	\end{tabular}}
\end{table*}

\begin{table}[t]
\centering
\footnotesize
\setlength{\tabcolsep}{4pt}
\renewcommand{\arraystretch}{1.08}
\caption{Performance comparison on the R2R dataset for Validation Unseen and Test Unseen splits.}
\label{tab:r2r_results}
\resizebox{\columnwidth}{!}{
\begin{tabular}{lcccc|cccc}
\toprule
 & \multicolumn{4}{c}{\textbf{Validation Unseen}} & \multicolumn{4}{c}{\textbf{Test Unseen}} \\
\cmidrule(lr){2-5} \cmidrule(lr){6-9}
\textbf{Model} & \textbf{TL}& \textbf{NE}& \textbf{SR} & \textbf{SPL} & \textbf{TL} & \textbf{NE}  & \textbf{SR} & \textbf{SPL}\\
\midrule
Seq2Seq \cite{anderson2018vision} & \textbf{8.39} & 7.81 & 22 & - & \textbf{8.13} & 7.85 & 20 & 18 \\
RecBERT \cite{hong2021vln} & 12.01 & 3.93 & 63 & 57 & 12.35 & 4.09 & 63 & 57 \\
HAMT \cite{chen2021history} & 11.87 & 3.65 & 65 & 59 & 12.65 & 4.11 & 63 & 58 \\
DUET \cite{chen2022think} & 13.94 & {3.31} & 72 & 60 & 14.73 & 3.65 & \textbf{69} & 59 \\
\textbf{DUET+SDB} & 13.49 & \textbf{3.26} & \textbf{72} & \textbf{62} & 14.40 & \textbf{3.58} & 70 & \textbf{61} \\
\bottomrule
\end{tabular}}
\end{table}

\subsection{Implementation Details}
\label{sec:impl}
\textbf{Primary controlled setting (DUET).}
Our main controlled comparisons are conducted with VLN-DUET, following its official training/evaluation protocol and hyperparameters.
All experiments are implemented in PyTorch and run on NVIDIA RTX 3090 GPUs (24GB).
We use ViT-B/16 panoramic features as in DUET, with maximum instruction length 200 and maximum action length 15.
We train with DAgger-style imitation learning (SPL expert policy) using AdamW with learning rate $1\times 10^{-5}$, batch size 6, for 200k iterations.
We set dropout to 0.5 and feature dropout to 0.4.

\textbf{SDB configuration (DUET setting).}
Unless otherwise stated, we use $K{=}3$ hypotheses (including the unshifted anchor) and a low-rank shift subspace of rank $r{=}32$.
Hypotheses are consolidated via a lightweight selector with soft weighting during training, and we optimize SDB with the auxiliary regularizer $\mathcal{L}_{\mathrm{sdb}}$ defined in Sec.~\ref{sec:method}.
Additional backbone-specific details for GOAT and NavGPT-2 follow the original papers and are deferred to the Appendix for completeness.

\subsection{Main Results}
We evaluate SDB on R2R, SOON, and REVERIE and compare against (i) our backbone baseline under {identical} settings (e.g., DUET vs.\ DUET+SDB), and (ii) published results reported under each method's original configuration.
For non-DUET methods in the tables, visual features, training strategies, and inference settings follow those used in the corresponding papers; we report them for reference, while our primary evidence comes from controlled ``+SDB'' improvements under matched settings.

\textbf{Results on REVERIE.}
Table~\ref{tab:reverie_results} shows that SDB improves goal-reaching performance in unseen environments under the same DUET training and evaluation protocol.
On {val\_unseen}, SDB increases SPL from 33.73 to 35.93, accompanied by higher SR (46.98$\rightarrow$50.81) and OSR (51.07$\rightarrow$54.25).
Similar gains are observed on {test\_unseen}, where SPL improves from 36.06 to 37.74 and SR increases from 52.51 to 54.92 (OSR: 56.71$\rightarrow$58.63).
While TL slightly increases on unseen splits (e.g., 21.59$\rightarrow$22.35 on {test\_unseen}), the higher SPL indicates that the policy achieves better success despite modestly longer trajectories, suggesting more reliable step-wise decision making and goal localization.

\textbf{Results on R2R.}
Table~\ref{tab:r2r_results} reports results on R2R under the standard validation unseen and test unseen splits.
SDB improves navigation efficiency while maintaining comparable or better success:
on val\_unseen, SPL improves (60$\rightarrow$62) with slightly shorter TL (13.94$\rightarrow$13.49) and reduced NE (3.31$\rightarrow$3.26), while SR remains unchanged (72);
on test\_unseen, SPL improves (59$\rightarrow$61) and SR increases (69$\rightarrow$70) with lower NE (3.65$\rightarrow$3.58).

\textbf{Results on SOON.}
Table~\ref{tab:soon_results} reports results on SOON.
Compared with DUET, SDB improves SPL on unseen splits (22.58$\rightarrow$25.00 on val\_unseen; 21.42$\rightarrow$23.84 on test\_unseen), together with substantially shorter trajectories on val\_unseen (TL: 36.20$\rightarrow$30.95), suggesting reduced dithering and more efficient long-horizon decision making.
Meanwhile, we observe a mild degradation in reachability-related metrics on SOON (e.g., SR/OSR decrease on unseen splits), indicating an efficiency--reachability trade-off in this more demanding long-horizon setting.
We hypothesize this to SDB's reliability-driven consolidation encouraging earlier commitment/termination under high uncertainty, which can reduce unnecessary exploration but may also affect stopping behavior in some cases.

\begin{table}[t]
\centering
\footnotesize
\setlength{\tabcolsep}{4pt}
\renewcommand{\arraystretch}{1.08}
\caption{Performance comparison on the SOON dataset for Validation Unseen and Test Unseen splits.}
\label{tab:soon_results}
\resizebox{\columnwidth}{!}{
\begin{tabular}{lcccc|cccc}
\toprule
 & \multicolumn{4}{c}{\textbf{Validation Unseen}} & \multicolumn{4}{c}{\textbf{Test Unseen}} \\
\cmidrule(lr){2-5} \cmidrule(lr){6-9}
\textbf{Model} & \textbf{TL}& \textbf{OSR}& \textbf{SR} & \textbf{SPL} & \textbf{TL} & \textbf{OSR}  & \textbf{SR} & \textbf{SPL}\\
\midrule
GBE~\cite{zhu2021soon}  & \textbf{28.96} & 28.54 & 19.52 & 13.34
                       & \textbf{27.88} & 21.45 & 12.90 & 9.23 \\
DUET~\cite{chen2022think} & 36.20 & \textbf{50.91} & \textbf{36.28} & 22.58
                         & 41.83 & \textbf{43.00} & \textbf{33.44} & 21.42 \\
\textbf{DUET+SDB}          & 30.95 & 47.20 & 35.84 & \textbf{25.00}
                         & 36.58 & 39.29 & 33.00 & \textbf{23.84} \\
\bottomrule
\end{tabular}}
\end{table}

\begin{table}[t]
\centering
\footnotesize
\setlength{\tabcolsep}{4pt}
\renewcommand{\arraystretch}{1.08}
\caption{Effect of hypothesis number $K$ on REVERIE, evaluated on val\_seen and val\_unseen.}
\label{tab:diversity}
\resizebox{\columnwidth}{!}{
\begin{tabular}{lcccccccc}
\toprule
 & \multicolumn{4}{c}{\textbf{Validation Seen}} & \multicolumn{4}{c}{\textbf{Validation Unseen}} \\
\cmidrule(lr){2-5} \cmidrule(lr){6-9}
\textbf{$K$} & TL & OSR & SR & SPL & TL & OSR & SR & SPL \\
\midrule
0 & 13.86 & 73.86 & 71.75 & 63.94 & 22.11 & 51.07 & 46.98 & 33.73 \\
2 & 13.28 & 72.94 & 71.54 & 64.65 & 21.46 & 53.08 & 49.99 & 34.76 \\
3 & 13.83 & 70.56 & 69.50 & 62.07 & 23.22 & 54.25 & 50.81 & 35.93 \\
5 & 13.64 & 70.63 & 68.52 & 60.09 & 21.76 & 53.56 & 48.85 & 34.90 \\
\bottomrule
\end{tabular}}
\end{table}

\subsection{Step-wise Planning Change Rate (SPCR) Analysis}
\label{sec:spcr}
To further analyze decision adaptability {within an LLM-style VLN agent}, we compute Step-wise Planning Change Rate (SPCR) on NavGPT-2, which exposes explicit language thoughts/plans at each navigation step.
We extract the decoded step-wise plan text at step $t$ (after removing fixed prompt templates) and compute token-level edit distance using whitespace tokenization.
\begin{equation}
\mathrm{SPCR}_t = \frac{\mathrm{EditDistance}(T_{t+1}, T_t)}{\max(|T_{t+1}|, |T_t|, 1)},
\end{equation}
where $T_t$ denotes the extracted plan token sequence at navigation step $t$, and $\mathrm{EditDistance}(\cdot,\cdot)$ is the Levenshtein distance.

\begin{figure}[t]
    \centering
    \includegraphics[width=0.48\textwidth]{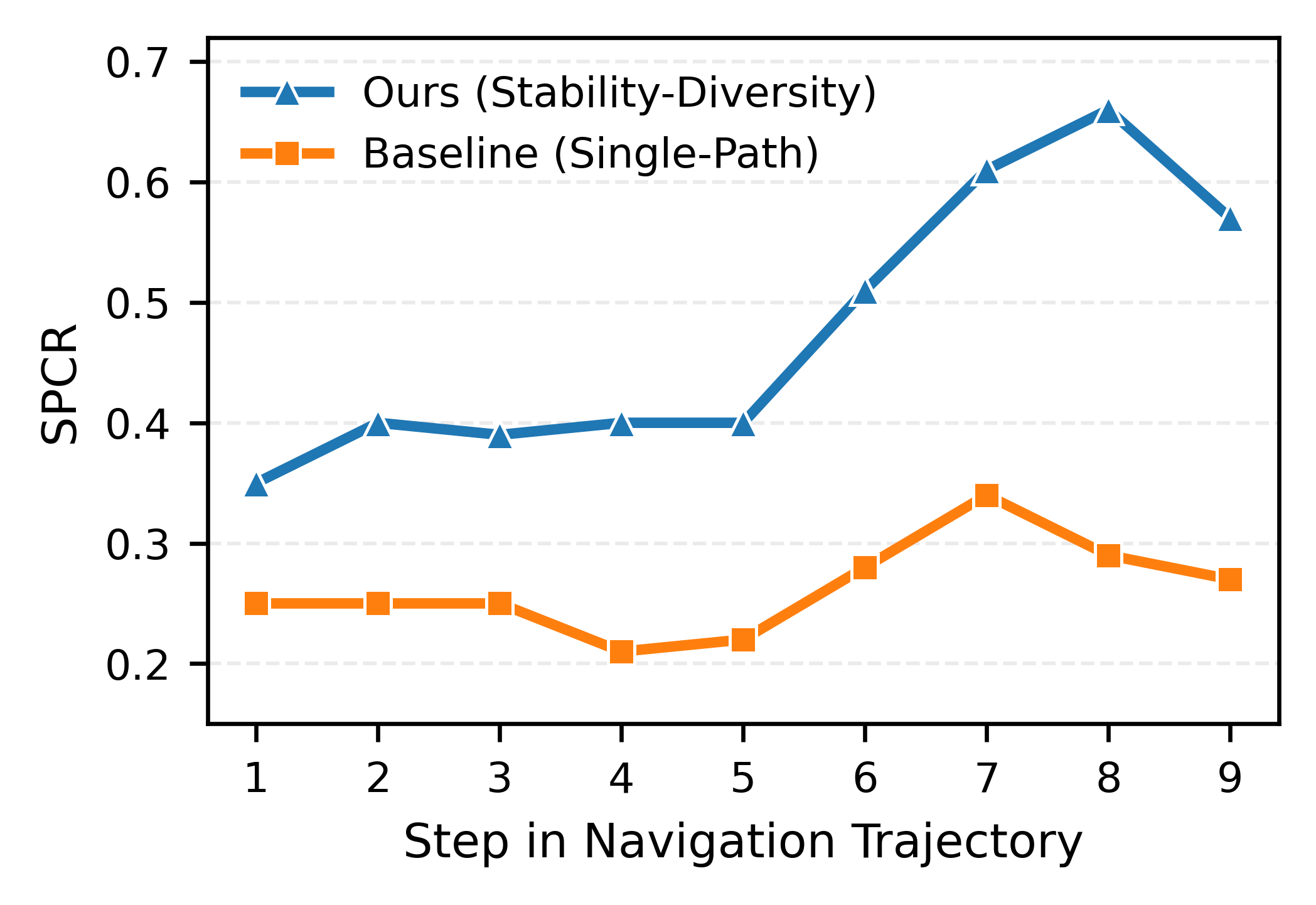}
    \caption{Step-wise Planning Change Rate (SPCR) on NavGPT-2, averaged over the first 9 navigation steps. SDB yields consistently higher change rates, indicating more dynamic and context-sensitive plan refinement.}
    \label{fig:spcr_curve}
\end{figure}

Higher SPCR indicates larger step-to-step plan revisions.
Across the first 9 navigation steps, our SDB-enhanced NavGPT-2 achieves a mean SPCR of 0.48 (std 0.11), higher than the single-path baseline's 0.26 (std 0.04).
As shown in Fig.~\ref{fig:spcr_curve}, our SPCR stays around 0.35--0.40 in early steps (1--5) and rises after step 6 (peaking around step 8), suggesting more substantial plan refinement as the trajectory evolves.

\subsection{Ablation Study}

\noindent\textbf{Effect of hypothesis number.}
Table~\ref{tab:diversity} studies the impact of the hypothesis number $K$ on {val\_seen} and {val\_unseen}, where $K{=}0$ corresponds to the DUET baseline without SDB.
On {val\_unseen}, increasing $K$ generally improves generalization: compared to $K{=}0$ (SPL 33.73, SR 46.98), $K{=}2$ raises SPL to 34.76 and SR to 49.99, and the best performance is achieved at $K{=}3$ with SPL 35.93 and SR 50.81 (OSR 54.25).
Further increasing to $K{=}5$ does not bring additional gains on unseen split (SPL 34.90, SR 48.85), suggesting diminishing returns when expanding too many hypotheses.
On {val\_seen}, a smaller $K$ is preferable: $K{=}2$ attains the best SPL (64.65), while larger $K$ reduces SR/SPL (e.g., 69.50/62.07 at $K{=}3$ and 68.52/60.09 at $K{=}5$), reflecting a stability--exploration trade-off on seen environments.
Balancing unseen performance and the approximately linear overhead in $K$, we adopt $K{=}3$ by default.


\begin{table}[t]
\centering
\footnotesize
\setlength{\tabcolsep}{4pt}
\renewcommand{\arraystretch}{1.08}
\caption{
Ablation of the Diversity Expansion Module (DEM) and the Stability Selection Module (SSM) on R2R val\_seen.
``HSG'' denotes our structured hypothesis generation; ``Noise'' denotes random-noise expansion; ``Stable'' denotes our stability-aware selection; ``Rand'' denotes random selection.
}
\label{tab:ablation}
\resizebox{\columnwidth}{!}{
\begin{tabular}{llccccc}
\toprule
\textbf{DEM} & \textbf{SSM} & \textbf{TL} & \textbf{NE}& \textbf{OSR}& \textbf{SR} & \textbf{SPL}\\
\midrule
HSG   & Rand   & \textbf{11.91} & 3.18 & 77 & 71 & 64 \\
Noise & Stable & 12.64 & 3.00 & 79 & 72 & 64 \\
Noise & Rand   & 12.51 & 3.00 & 80 & 72 & 64 \\
HSG   & Stable & {12.26} & \textbf{2.92} & \textbf{81} & \textbf{75} & \textbf{67} \\
\bottomrule
\end{tabular}}
\end{table}

\noindent\textbf{Ablation on DEM and SSM (NavGPT-2 on R2R).}
We conduct this ablation on NavGPT-2 under the R2R val\_seen split to disentangle the contributions of structured hypothesis expansion (DEM) and stability-aware consolidation (SSM).
Specifically, we compare structured expansion (HSG) against noise-based expansion, and stability-aware selection against random selection, yielding four combinations in Table~\ref{tab:ablation}.
This design tests whether the proposed expand--select procedure improves decision quality through plausible hypothesis generation and reliable consolidation, rather than through stochastic perturbations or chance selection.

\begin{figure*}[t]
    \centering
    \includegraphics[width=\textwidth]{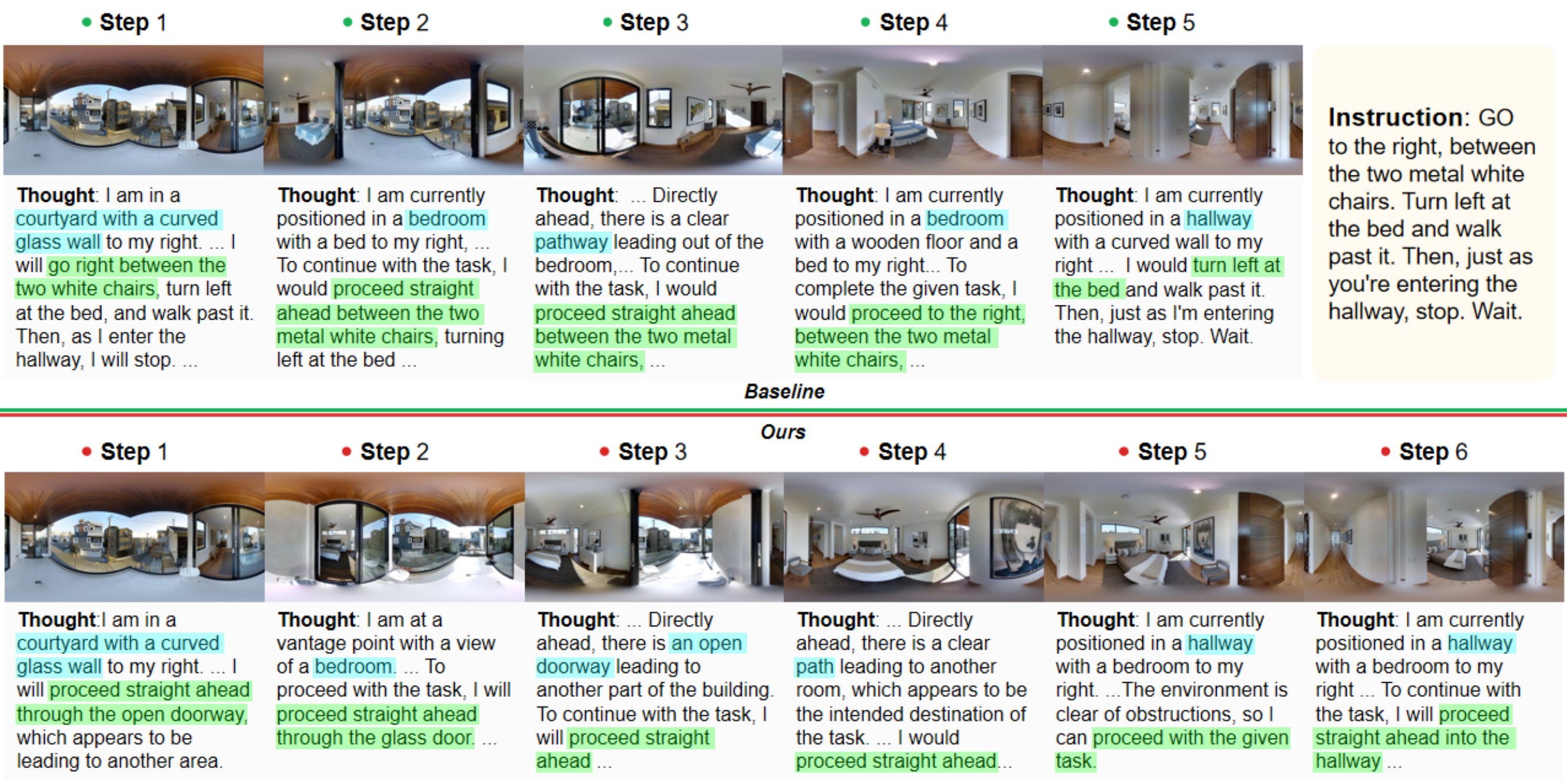}
    \caption{
        \textbf{Step-wise reasoning details.}
        Each column shows the agent's observation and the corresponding selected thought at each navigation step.
        \textbf{Top:} Baseline (NavGPT-2).
        \textbf{Bottom:} NavGPT-2 + SDB (ours).
        Step indices and colors correspond to the top-down view in Figure~\ref{fig:topdown_traj}.
        Compared with the baseline, SDB leads to more instruction-consistent decisions and more interpretable step-wise reasoning.
    }
    \label{fig:stepwise_detail}
\end{figure*}

\begin{figure}[t]
    \centering
    \includegraphics[width=\columnwidth]{\detokenize{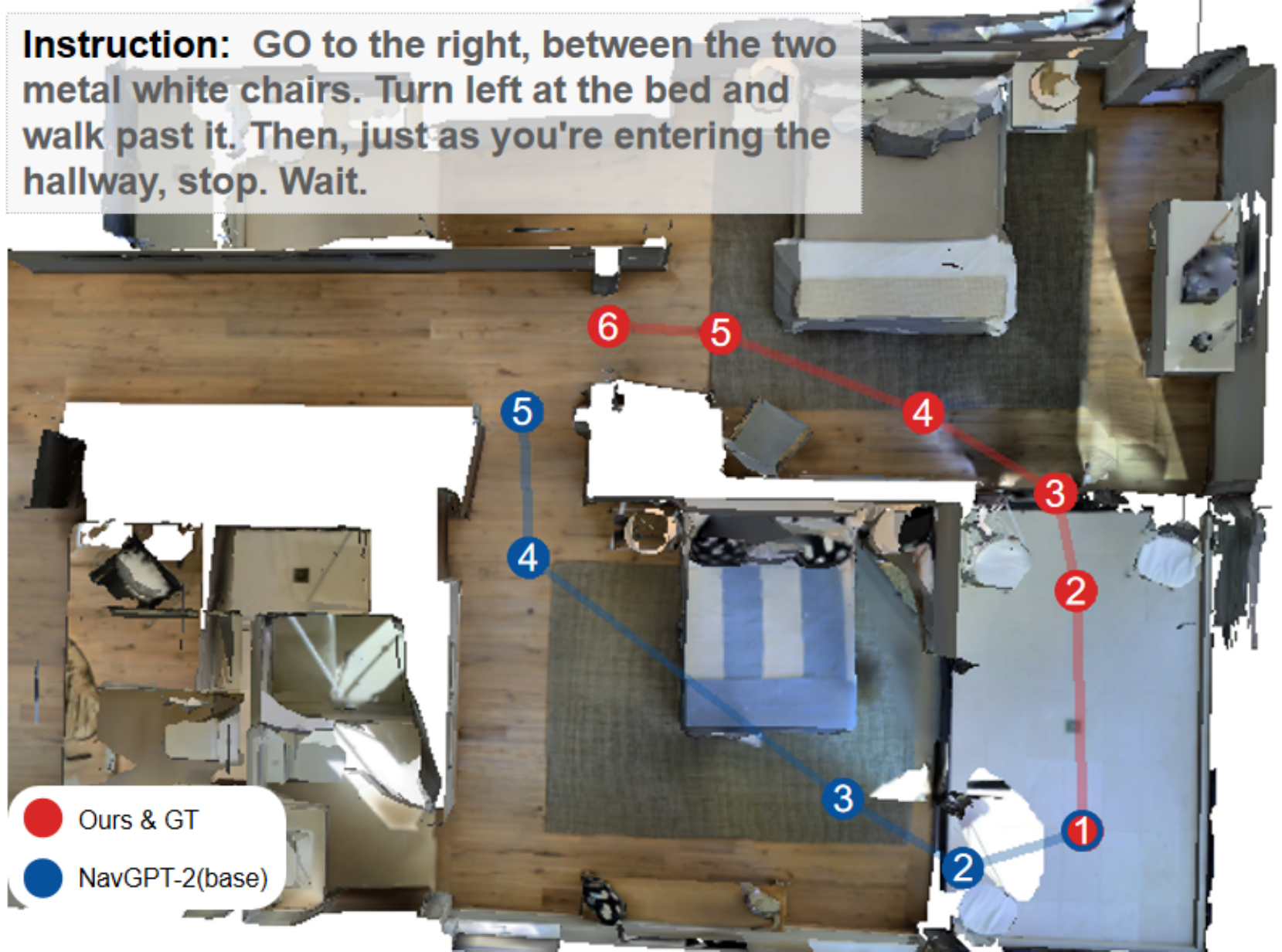}}
    \caption{
        \textbf{Top-down trajectory comparison} between NavGPT-2 + SDB (red) and the baseline (blue) under the same instruction.
        The SDB-enhanced agent follows the instruction more closely and reaches the goal, whereas the baseline deviates early.
        Step colors correspond to Figure~\ref{fig:stepwise_detail}.
    }
    \label{fig:topdown_traj}
\end{figure}

\noindent\textbf{Backbone generality across datasets.}
To verify that SDB is not tied to a specific backbone, we plug it into two substantially different VLN backbones and evaluate on their standard benchmarks (Table~\ref{tab:sdb_backbone_generality}).







\begin{table}[t]
\centering
\caption{Backbone generality of SDB across datasets.}
\label{tab:sdb_backbone_generality}
\footnotesize
\setlength{\tabcolsep}{4pt}
\renewcommand{\arraystretch}{1.08}

\textbf{(A) REVERIE / VLN-GOAT}\vspace{1mm}

\resizebox{\columnwidth}{!}{%
\begin{tabular}{lcccc}
\toprule
Methods & \textbf{SR}& \textbf{SPL}& \textbf{RGS}& \textbf{RGSPL}\\
\midrule
\multicolumn{5}{l}{\textbf{Val Seen}} \\
VLN-GOAT (base)   & 78.64 & 71.40 & 63.74 & 57.85 \\
VLN-GOAT + SDB    & \textbf{81.87} & \textbf{74.67} & \textbf{65.21} & \textbf{59.36} \\
\midrule
\multicolumn{5}{l}{\textbf{Val Unseen}} \\
VLN-GOAT (base)   & 53.37 & 36.70 & 38.43 & 26.09 \\
VLN-GOAT + SDB    & \textbf{54.42} & \textbf{39.00} & \textbf{39.11} & \textbf{27.95} \\
\bottomrule
\end{tabular}}

\vspace{1.5mm}

\textbf{(B) R2R / NavGPT-2}\vspace{1mm}

\resizebox{\columnwidth}{!}{%
\begin{tabular}{lccccc}
\toprule
Methods & \textbf{TL}& \textbf{NE}& \textbf{OSR}& \textbf{SR}& \textbf{SPL}\\
\midrule
\multicolumn{6}{l}{\textbf{Val Seen}} \\
NavGPT-2 (base)   & 12.44 & 2.97 & 80 & 73 & 65 \\
NavGPT-2 + SDB    & \textbf{12.26} & \textbf{2.92} & \textbf{81} & \textbf{75} & \textbf{67} \\
\midrule
\multicolumn{6}{l}{\textbf{Val Unseen}} \\
NavGPT-2 (base)   & 12.81 & 3.33 & 79 & 70 & 59 \\
NavGPT-2 + SDB    & 12.86 & \textbf{3.30} & \textbf{79} & \textbf{71} & \textbf{60} \\
\bottomrule
\end{tabular}}
\end{table}

\noindent\textbf{Robustness analysis.}
We evaluate robustness to training randomness by repeating DUET+SDB with three different random seeds on REVERIE; the {val\_unseen} SR/SPL improvements are consistent across runs with small variance.
We further ablate the reliability cues (A/C/S) and the stability regularizer; removing any component reduces {val\_unseen} SR/SPL, and removing the stability cue drops performance close to the DUET baseline.

\subsection{Additional Analysis}
\label{sec:additional_analysis}

This subsection provides a brief interpretation of the observed gains of SDB across the preceding studies.
Taken together, the ablations in Table~\ref{tab:ablation} and Table~\ref{tab:diversity} suggest that improvements are not driven by indiscriminate perturbation or simply increasing the number of hypotheses, but by generating plausible alternatives and consolidating them in a stable manner.
This interpretation is consistent with the SOON results in Table~\ref{tab:soon_results}, where efficiency-related improvements can appear without uniformly increasing all success-related metrics under higher uncertainty.
For VLM-based agents, the higher SPCR in Fig.~\ref{fig:spcr_curve} indicates more frequent plan updates, which complements the above view that SDB supports decision updates conditioned on new observations rather than rigid early commitment.
Finally, the consistent gains across backbones and datasets in Table~\ref{tab:sdb_backbone_generality} suggest that SDB improves a generally applicable sequential decision procedure, instead of relying on a specific model architecture.
\subsection{Qualitative Results}
\label{sec:qualitative}
Figure~\ref{fig:topdown_traj} provides a qualitative comparison on an instruction sampled from the R2R validation set, showing top-down trajectories of the baseline and the SDB-enhanced agent, with step-wise observations and selected decisions visualized in Figure~\ref{fig:stepwise_detail}~\cite{li2025regnav}.
At each step, SDB maintains multiple candidate hypotheses and applies stability-aware selection to determine the final action; for clarity, we visualize only the selected thought per step, while representative candidates (top-ranked by the controller) are deferred to the Appendix~\cite{qi2026patchcue,fu2024understanding}.

We use NavGPT-2 for qualitative visualization because it exposes explicit thought traces, making the effect of SDB on decision stability and instruction grounding easier to inspect~\cite{wu2025event}.
Importantly, SDB itself is backbone-agnostic and our quantitative gains are demonstrated under both specialist VLN backbones (e.g., DUET/GOAT) and an LLM-style backbone (NavGPT-2)~\cite{li2025regnav,li2024camera}.

More concretely, Fig.~\ref{fig:topdown_traj} highlights a critical branching point (around Step~2), where the baseline commits early to an off-route branch.
In contrast, SDB avoids premature commitment and stays closer to the instruction-aligned corridor until the visual evidence becomes discriminative, resulting in a trajectory that better matches the intended route~\cite{liu2024semantic,liu2025mind}.
This difference is further reflected in Fig.~\ref{fig:stepwise_detail}: when the instruction admits multiple plausible next actions, the baseline's thoughts tend to rely on a local cue and produce a mismatched action, whereas SDB maintains alternative candidates and selects a hypothesis that remains consistent with both the instruction and newly observed context~\cite{wu2024cr,fu2019recognition,qi2026patchcue}.
These qualitative behaviors align with our mechanism analysis: structured hypothesis expansion preserves plausible interpretations under ambiguity, and stability-aware selection prevents uncoordinated switching or premature collapse to a single hypothesis~\cite{wang2023closing,fu2023and,fu2023breaking}.

Overall, the visualization provides intuitive evidence that SDB improves step-wise grounding under ambiguity by coordinating hypothesis exploration and stable selection.

\section{Conclusion}

This paper presented Stability--Diversity Balance (SDB), a plug-in decision framework for vision--language navigation that explicitly addresses the stability--diversity tension in self-improvement under standard action-label supervision.
SDB instantiates a structured $(1\!\rightarrow\!K\!\rightarrow\!1)$ expand--select operator: a Diversity Expansion stage generates multiple latent decision hypotheses from the same instruction--environment evidence, and a Stability Selection stage performs reliability-aware consolidation for training while enabling more stable hypothesis selection during execution.
SDB is modular and can be inserted between a backbone encoder and an action head without modifying the backbone architecture.
Across R2R, SOON, and REVERIE, and under multiple representative backbones, SDB generally improves navigation performance and path efficiency, with particularly clear efficiency gains in long-horizon and ambiguous scenarios where premature commitment or uncoordinated diversity can destabilize learning.
Motivated by the observed sweet spot in $K$, future work includes adapting the hypothesis number online, as well as enriching reliability cues to further improve robustness under stronger distribution shifts.

\vspace{-33pt}
\begin{IEEEbiography}[{\includegraphics[width=1in,height=1.25in,clip,keepaspectratio]{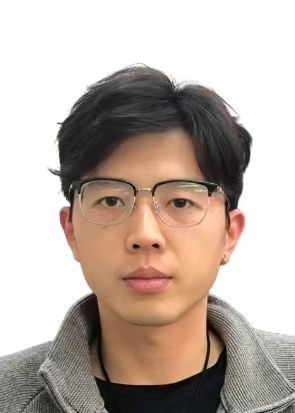}}]{Zhen Liu}
is currently a Ph.D. student at the National Key Laboratory of Human-Machine Hybrid Augmented Intelligence, Xi'an Jiaotong University. His research focuses on vision-and-language navigation and embodied intelligence.
\end{IEEEbiography}
\vspace{-33pt}
\begin{IEEEbiography}[{\includegraphics[width=1in,height=1.25in,clip,keepaspectratio]{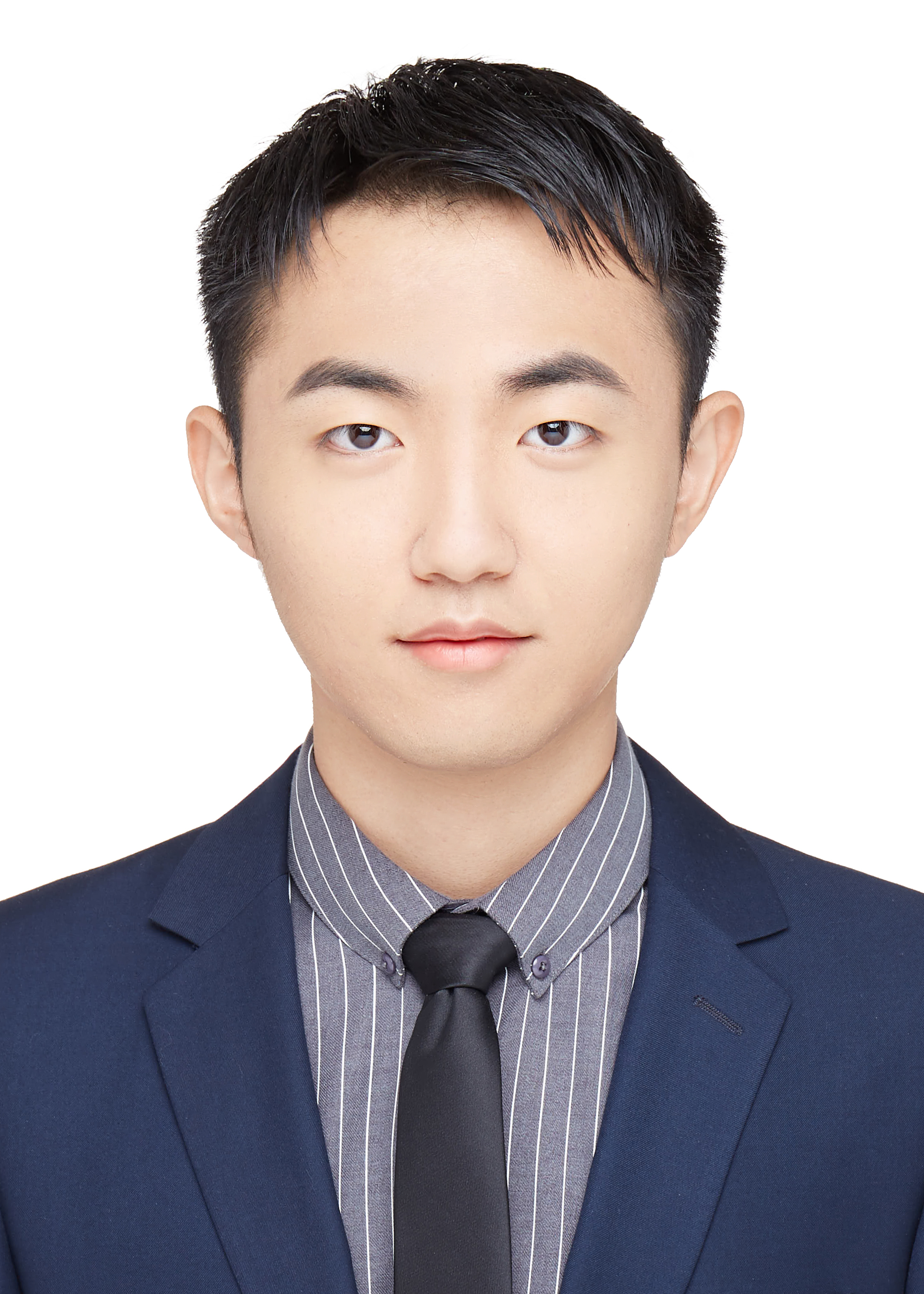}}]{Yuhan Liu}
 received the B.S. degree from Department of Computer Science and Technology (Honors Science Program) and satisfied the requirements of the Honors Youth Program in Xi’an Jiaotong University in 2021. He is currently pursuing Ph.D. degree with the Institute of Artificial Intelligence and Robotics, Xi’an Jiaotong University. His research interests include computer vision, representation learning and image matching.
\end{IEEEbiography}

\vspace{-33pt}
\begin{IEEEbiography}[{\includegraphics[width=1in,height=1.25in,clip,keepaspectratio]{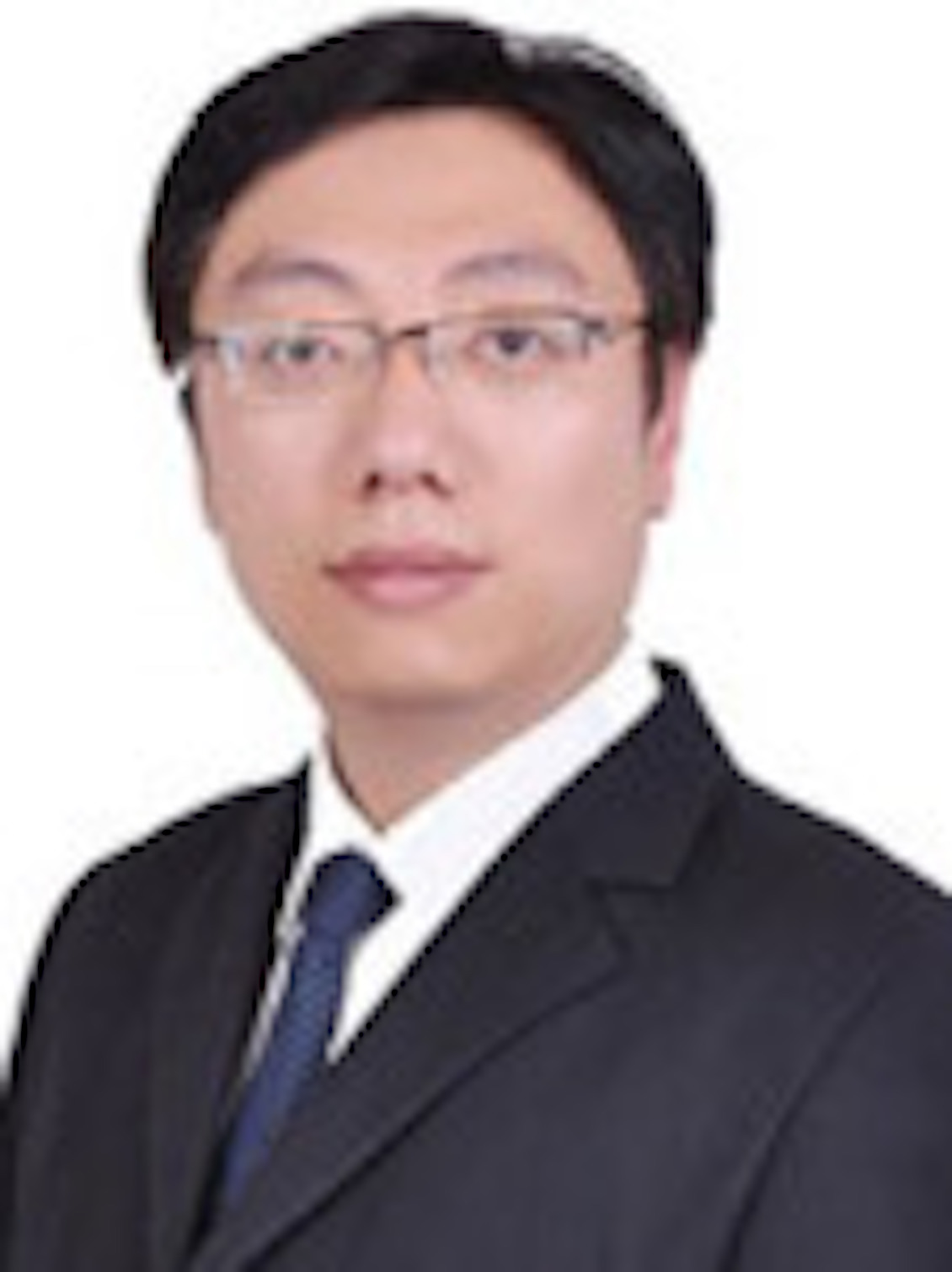}}]{Jinjun Wang} received the B.E. and M.E. degrees from the Huazhong University of Science and Technology, China, in 2000 and 2003, respectively. He received his Ph.D. degree from Nanyang Technological University, Singapore, in 2006. From 2006 to 2009, he was a Research Scientist with NEC Laboratories America, Inc., San Jose, CA, USA. From 2010 to 2013, he was a Senior Research Scientist with Epson Research and Development, Inc., Cupertino, CA, USA. He is currently a Professor at Xi’an Jiaotong University, Xi’an, China.
His research interests include pattern classification, image/video enhancement and editing, content-based image/video annotation and retrieval, and semantic
event detection, etc.
\end{IEEEbiography}

\vspace{-33pt}
\begin{IEEEbiography}[{\includegraphics[width=1in,height=1.25in,clip,keepaspectratio]{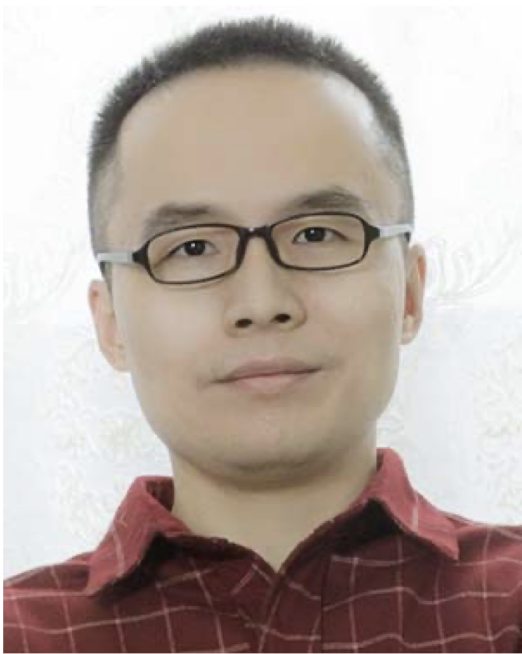}}]{Jianyi Liu}
received the B.Eng. and the M.Eng. degrees in information and communication engineering and the Ph.D. degree in control science and engineering from Xi'an Jiaotong University, Xi'an, China, in 1998, 2001, and 2009, respectively. He is currently an Associate Professor with the College of Artificial Intelligence, Xi'an Jiaotong University. His research interests include image processing, pattern recognition, and machine learning.
\end{IEEEbiography}

\vspace{-33pt}
\begin{IEEEbiography}[{\includegraphics[width=1in,height=1.25in,clip,keepaspectratio]{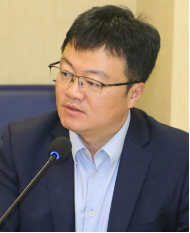}}]{Wei Song}
is a professor at the Department of Computer Science and Technology of North China University of Technology. He received his B. Eng. Degree in Software Engineering from Northeastern University, China, in 2005, and his M. Eng. and PhD. Eng. in the department of Multimedia from Dongguk University, Seoul, Korea, in 2008 and 2013, respectively. His current research interests are focused on LiDAR, object recognition, semantic segmentation, intelligent transportation system, SLAM, and parallel computation. He has published over 100 international conference and journal papers.
\end{IEEEbiography}

\vspace{-33pt}
\begin{IEEEbiography}[{\includegraphics[width=1in,height=1.25in,clip,keepaspectratio]{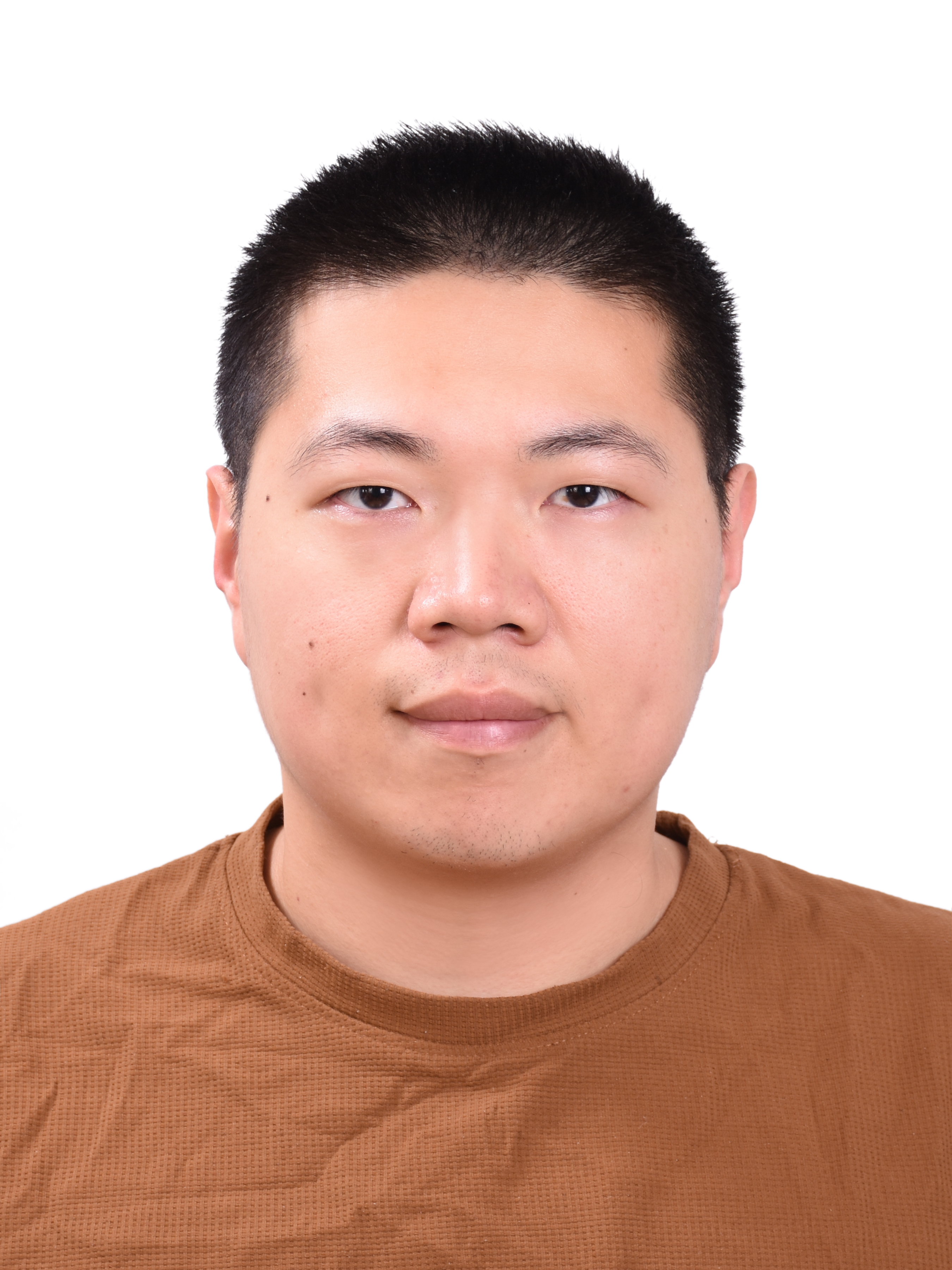}}]{Jingwen Fu}
 graduated from the School of Artificial Intelligence at Xi'an Jiaotong University. During his doctoral studies, he was selected for the joint PhD program between Xi'an Jiaotong University and Microsoft Research Asia (MSRA).  Additionally, he spent one year as a visiting scholar at RIKEN in Japan. Dr. Fu has long been dedicated to fundamental research in artificial intelligence. His current core research interests focus on the interpretability of Large Language Models, the intersection of LLMs and linguistics, composition generalization, and multi-agent systems. To date, he has published over 10 papers in top-tier international academic conferences and journals in the field of AI and has been granted several national invention patents.
\end{IEEEbiography}

\end{document}